\documentclass{article}

\usepackage[english]{babel}
\usepackage[utf8x]{inputenc}
\usepackage{amsmath}
\usepackage{lineno,hyperref}
\usepackage{algpseudocode}
\usepackage{natbib}
 
\addto\captionsenglish{}

\usepackage{program}
\usepackage{multirow}
\modulolinenumbers[5]

\newcommand{\mathCfg}{
\abovedisplayskip=0pt plus 0pt minus 0pt
\abovedisplayshortskip=0pt plus 3pt
\belowdisplayskip=0pt plus 0pt minus 0pt
\belowdisplayshortskip=7pt plus 3pt minus 4pt
}

\makeatletter
\def\THEN{\@marginspace\untab\keyword{then}\ \tab}
\def\ELSE{\@marginspace\untab\keyword{else}\ \tab}%
\def\FI{\@marginspace\untab}
\makeatother
 
\begin{document}

\newtheorem{algorithm}{Algorithm}[section]

\title{A hybrid optimization procedure for solving a tire curing scheduling problem}

\author{Joaqu\'in Vel\'azquez 
\and H\'ector Cancela \and Pedro Pi\~neyro\\
Departamento de Investigación Operativa, Instituto de Computaci\'on \\ Facultad de Ingenier\'ia, Universidad de la Rep\'ublica\\  Julio Herrera y Reissig 565, 11300\\ Montevideo, Uruguay}

\maketitle

\begin{abstract}
This paper addresses a lot-sizing and scheduling problem variant arising from the study of the curing process of a tire factory. The aim is to find the minimum makespan needed for producing enough tires to meet the demand requirements on time, considering the availability and compatibility of different resources involved. To solve this problem, we suggest a hybrid approach that consists in first applying a  heuristic to obtain an estimated value of the makespan and then solving a mathematical model to determine the minimum value. We note that the size of the model (number of variables and constraints) depends significantly on the estimated makespan. Extensive numerical experiments over different instances based on real data are presented to evaluate the effectiveness of the hybrid procedure proposed. From the results obtained we can note that the hybrid approach is able to achieve the optimal makespan for many of the instances, even large ones, since the results provided by the heuristic allow to reduce significantly the size of the mathematical model.
\end{abstract}

\section{Introduction}\label{intr}
The tire manufacturing process is usually decomposed into three steps: materials mixing, tire building and tire curing or  vulcanization. In the first step, different materials such as natural or synthetic rubber, carbon black, sulfur, oils and other chemicals are mixed to prepare the rubber sheets. Then, the rubber sheets along with fabric and wires are used to build the uncured or green tires. Finally, in the tire curing step, the green tires are placed into curing machines known as heaters, to apply the needed levels of heat and pressure to give the tires the required functional characteristics. Molds are used in the heaters to give the tires their shape and tread pattern. The curing process has a large impact in the manufacturing efficiency, as it employs expensive equipment and consumes a large amount of energy. Adequately planning the tire curing process to meet the demand requirements on time is in general a complex task. It is necessary to consider the number of available heaters and molds as well as the mold-mold and mold-heater allowed combinations. In addition, mold configuration and tire curing times must be considered in the planning process. 

Particular cases of tire curing scheduling problems are studied in the seminal works of \citet{Gorenstein1970} and \citet{Lasdon1971}. More recently, \citet{Degraeve1997} address the tire curing scheduling problem for the tire manufacturer Bridgestone/Firestone Off-The-Road, in which the tires have different priority and the number of compatible molds that can be stacked into one heater is limited. A column generation based procedure is proposed for solving the problem. Later, in \citet{Degraeve1998}  a scheduling software for this problem is presented. \citet{Jans2004} also consider the column generation approach for the tire curing scheduling problem of the Solideal manufacturer. The problem is modeled as an extension of the general Discrete Lot-Sizing and Scheduling Problem (DLSP) that involves start-up times, mold-heater compatibility, capacity constraints and backlogging. The objective is to determine the tire curing plan that minimizes the sum of the costs involved. For details about the DLSP we refer the readers to \citet{Fleischmann1990}, \citet{Salomon1991} and \citet{Bruggemann2000} and to \citet{Copil2017} for a recent survey. 
Several works consider the two-stage production system of building and curing of tires. \citet{Tabucanon1991} tackle both problems separately, considering first the minimization of the production costs of the green tires and then the curing scheduling problem of them. \citet{Kuiteng2006} and \citet{Oulamara2009} formulate the integrated problem of building and curing tires as a Flow Shop Scheduling Problem (FSSP) considering a single batching machine and task compatibilities, with the objective of minimizing the makespan. Three different heuristic procedures along with a simple lower bound procedure are suggested and evaluated for the problem. \citet{Bellanger2009} extend the two works mentioned above by considering multiple machines at each stage. A Polynomial Time Approximation Scheme (PTAS) algorithm is suggested for the particular case in which tasks have the same processing time on the first stage. \citet{Safari2017} also consider the integrated problem of building and curing tires by means of a Hybrid Flow-Shop Scheduling Problem (HFS) model with resource constraints. The procedure proposed for solving the problem is based on hybridizing the Genetic Algorithms (GA) and Variable Neighborhood Search (VNS) metaheuristics. For details about the FSSP we refer to \citet{Johnson1954} and \citet{Graham1979}. We also refer to \citet{Brucker2007}  and \citet{Ribas2010} for recent surveys on FSSP and HFS, respectively. 
\citet{Mukhopadhyay2005} and \citet{Karabicak2017} study the implementation in the production line of a tire manufacturer of the Kanban and Kaizen systems, respectively. \citet{Yu2011,Yu2012,Yu2013} provide heuristic procedures and simulation analysis for tire curing scheduling problems under particular specifications. \citet{Arslankaya2016} study the production process of trial tires by means of simulation techniques aiming to improve the productivity of a tire manufacturer. Finally, \citet{Kim2017} consider the integrated problem of materials mixing and building of tires. A Particle Swarm Optimization (PSO) based procedure is suggested for solving the problem.

The problem tackled in this paper has been defined to model the tire curing process at FunsaCoop (\url{http://funsa.uy}), an Uruguayan tire manufacturer. FunsaCoop is a cooperative organization producing tires for both the internal market and for exportation to other countries in the region. Some characteristics of FunsaCoop tire curing process include mold setup and removal times, shared parts among molds, and heaters holding up to two molds simultaneously. While these features are usual in real life situations in a tire factory (not just at FunsaCoop), as far as we know they have not been modelled in detail in the existing literature. 

In a previous work \citep{CPV2018}, we introduced a Mixed Integer Linear Programming (MILP) formulation for the tire curing problem at FunsaCoop. That model was employed for solving small instances but it was not suitable for finding exact solutions for larger instances, as the MILP solver run out of time or out of memory before finishing. These results motivated the study of other solution approaches. In the current paper, we suggest an hybrid approach, consisting in two phases: in the first phase a heuristic solution procedure is used to determine a  feasible solution of good quality; in the second phase, the makespan of this solution is used as a planning horizon parameter to instantiate a MILP formulation, which is then tackled using a solver. This hybrid optimization procedure is evaluated over scenarios  comprising instances of different sizes generated from real data given by FunsaCoop.

The remainder of the paper is organized as follows. In Section \ref{form} we present the description of the tire curing problem under study and a MILP formulation. In Section \ref{hybrid} we describe the hybrid approach proposed for solving the problem. Section \ref{experiments} provides the computational results obtained by applying the different solving approaches to a number of test scenarios inspired from real data. Section~\ref{conclusions} presents the general conclusions of the paper and possible directions for future work.

\section{Problem formulation}\label{form}
	
As mentioned before, FunsaCoop
manufactures different type of tires for the regional market. The production is pulled by the demand requirements, which are known in advance and must be satisfied within a given time horizon. Thus, the production of tires at FunsaCoop can be classified as a make-to-order production system.

The study presented in this paper focuses on the scheduling of the curing process. The heaters are set up with the molds corresponding to the type of tires to be produced. Every heater can hold up to two molds at once. Some molds are composed by special parts also known as \emph{molars}. Parts can not be shared while they are in use, and their number is limited. During the curing process, the green tires are exposed to high pressures and temperatures in order to acquire their final characteristics. As the curing time depends on the tire type, the pair of molds within a heater must be either identical, or belonging to a set of compatible molds for tires having similar curing times. In addition to the molds compatibility, only some combinations between molds and heaters are allowed. Therefore, the number of different tires of the same type that can be processed simultaneously is limited, since it depends on the number of available heaters, molds and  parts. The workers perform a number of tasks when molds are swapped in a heater, such as placement, cleanup, machinery warming and removal. The times of these tasks define the setup and removal times, which reduce the duration of a working period when they are carried out.


In FunsaCoop the heaters are powered by a steam boiler, which is the most expensive operative component of the company. The survey at FunsaCoop allowed to determine that makespan minimization is the most suitable objective function for the tire curing scheduling problem under study. In addition to the reduction of production and operative costs in terms of the steam boiler, makespan minimization decreases the $CO_2$ emissions and improves the delivery times in a make-to-order setting. In order to determine high quality solutions in terms of makespan, it is necessary to balance the trade-off between curing and configuration times. The curing times are reduced if there are as many molds as possible processed in parallel. On the other hand, to reduce the setup and removal times it is necessary to reduce as much as possible the number of mold changes in the heaters.
%
%

All the aforementioned characteristics of the tire curing scheduling problem at FunsaCoop were considered in order to develop the mathematical formulation presented below. The model, which can be considered as a variant of the DLSP, is a revised version of an initial MILP formulation introduced in \citet{CPV2018} for the problem.
	
	
	Sets:
	\mathCfg
	\begin{align*}	
	0 = & \text{ Empty mold identifier.} \\
	M  = & \text{ Molds set, indexes $i,j,l \in \{1...|M|\}$.} \\
	MExt = & \text{ } \{0\} \cup M = \text{ $M$ extended with the empty mold.} \\
	H  = & \text{ Heaters set, indexes $k, k' \in \{1...|H|\}$.} \\
	P = & \text{ Periods set, indexes $t \in \{1...|P|\}$.}  \\
    PExt = &  \text{ $P$ extended with period zero, indexes $t' \in \{0\} \cup P$.}\\
	Q  = & \text{ Parts set, indexes $q \in \{1...|Q|\}$.}\\
	C = & \text{ Set of compatible mold-heater pairs $(i, k)$.}\\
	CExt = &\text{ } C \cup \{ (0,k) \mid k \in H \} \text{, extended $C$ set, including the empty mold } \\
	& \text{ which is compatible with each heater.} \\
	MC = & \text{ Set of molds that can be processed together $(i, j)$  with  $i \leq j$.} \\
	MCExt = & \text{ $MC \cup \{(0,i) \mid i \in M \}$, including the empty mold which is compatible with all others.} 
	\end{align*}
    Parameters:
	\mathCfg
	\begin{align*}
    \text{THB}  = & \text{ Time horizon bound for production planning, with $\text{THB}=|P|$.} \\
	dm_{i}  =  & \text{ Demand of tires of type $i$ (i.e., tires produced with mold $i$).} \\
	tc_i     =  & \text{ Time needed to place mold $i$ in a heater (setup time).} \\
	tq_i     =  & \text{ Time needed to remove mold $i$ from a heater (removal time).} \\
	tv_{i,k} =  & \text{ Curing time of tires of mold type $i$ in heater $k$.}\\
	\phi      =  & \text{ Duration of working day.}\\
	nm_i    =  & \text{ Quantity of molds $i$ available.} \\
	np_{q} = & \text{ Quantity of parts of type $q$ available. } \\
	rq_{iq} = & \text{ Binary parameter that indicates if mold $i$ requires part $q$.}\\
	init_{ik} = & \text{ Initial quantity of molds $i$ used in heater $k$, with }init_{ik}\in\{0,1,2\}. 
\\
	\end{align*}
	Auxiliary sets:	
	\begin{align*}
	T  = & \text{ $\{ (i, j, k) \mid (i, j) \in MC \wedge \{(i,k),(j,k)\} \subseteq C  \wedge i \leq j   \}$.} \\
	&\text{ Compatible pairs of molds $(i,j)$ that can be processed in heater $k$.} \\
	T_i = & \text{ $\{ (j, k) \mid (i, j,k ) \in T \}$.} \\
	PC_q = & \text{ $\{i \mid rq_{iq} = 1, \forall i\in M \}$, with $q\in Q$, }  \text{set of molds that share part $q$.} \\
	TExt = & \text{ $\{(i,j,k) \mid (i,j) \in MCExt \wedge \{(i,k),(j,k)\} \subseteq CExt \wedge i \leq j \}$.} \\
	& \text{ Extended set $T$ including the empty mold.}\\
	TExt'_i = & \text{ $\{ (j, k) \mid (j,i,k) \in TExt \}$.} \\
	HExt_k = & \text{ $\{(i,j) \mid (i,j,k) \in TExt \}$.} 
\\
	\end{align*}

	Variables:	
	\mathCfg
	\begin{align*}
	x_{ikt}    = & \text{ Number of molds $i$ used in heater $k$ during period $t$, with $x_{ikt} \in \{0,1,2\}.$} \\
	y_{ikt}    = & \text{ Number of setups of mold $i$ in heater $k$ during period $t$ with $y_{ikt}\in\{0,1,2\}$.} \\
	y'_{lkt}   = & \text{ Number of molds $l$ used in last period and which must be removed at} \\
	& \text{heater $k$ at period $t$, with $y'_{lkt} \in \{0,1,2\}$.} \\
	z_{ijkt}   = & \text{ Binary variable that indicates if molds $i$ and $j$ are used in heater $k$} \\
	& \text{ during period $t$.}\\
	w_t        = & \text{ Binary variable that indicates if at period $t$ there exists some heater in use.} \\
	u_{ijkt} = & \text{ Number of tires cured using mold $i$ when assigned jointly } \\
	& \text{ with mold $j$  in heater $k$ during  period $t$.} \\
	prd_{it}   = & \text{ Number of tires cured using mold $i$ during period $t$.}
	\end{align*}
	
	The formulation is then as follows:
	\setcounter{equation}{0}
	\mathCfg
	\begin{align}
	& \text{min} && \sum_{ t \in P } w_t   \label{eq-1} \\
	& \text{s.t.} && w_t \leq w_{t-1}, \forall t \in P \backslash \{1\} \label{eq-2} \\
	&&& w_t \geq \frac{1}{2|H|} \sum_{(i,j,k)\in TExt} z_{ijkt}, \forall t\in P \label{eq-3} \\
	&&& \sum_{(i,j)\in HExt_k} z_{ijkt} \leq 1, \forall k \in H, \forall t\in P \label{eq-4} \\
	&&& u_{0jkt} \leq  \frac{( \phi - y_{jkt}.tc_j - \sum_{l \in M } y'_{lkt}.tq_l)} {tv_{jk}}, \forall (0,j,k) \in T, \forall t \in P\ \label{eq-5} \\
	&&& u_{ijkt} \leq \frac{( \phi - y_{ikt}.tc_i - y_{jkt}.tc_j - \sum_{l\in M} y'_{lkt}.tq_l)} {max( tv_{ik}, tv_{jk} )}, \forall (i,j,k) \in T, i < j, \forall t \in P\ \label{eq-6} \\
	&&& u_{iikt} \leq \frac{( \phi - y_{ikt}.tc_i - \sum_{l\in M} y'_{lkt}.tq_l)} { tv_{ik}}, \forall (i,i,k) \in T, \forall t \in P\ \label{eq-6bis} \\
	&&& u_{ijkt} \leq \frac{\phi}{max( tv_{ik}, tv_{jk} )}.z_{ijkt}, \forall (i,j,k)\in TExt, \forall t \in P \label{eq-7} \\
	&&& prd_{it} = \sum_{(j,k)\in T_i} u_{ijkt} + \sum_{ (j,k) \in TExt'_i } u_{jikt}, \forall i \in M, \forall t \in P \label{eq-8} \\
	&&&  \sum_{t \in P} prd_{it} \geq dm_i, \forall i \in M \label{eq-9} \\
	&&& x_{ikt} = \sum_{(j,k) \in T_i } z_{ijkt} + \sum_{(j,k) \in TExt'_i } z_{jikt}, \forall i \in M, \forall k \in H, \forall t \in P\label{eq-10} \\
	&&& \sum_{ (i,k)\in C } x_{ikt} \leq nm_i, \forall i\in M, \forall t \in P \label{eq-11} \\
	&&& \sum_{i \in PC_q} x_{ikt} \leq np_q, \forall q\in Q, \forall k\in H, \forall t\in P\label{eq-12} \\
	&&& x_{ik0} = init_{i,k}, \forall i\in MExt, \forall k \in H \label{eq-13} \\
	&&& y_{ikt} \geq x_{ikt} - x_{ik(t-1)}, \forall i \in M, \forall k \in H , \forall t \in P \label{eq-14} \\ 
	&&& y'_{ikt} \geq x_{ik(t-1)} - x_{ikt}, \forall i \in M, \forall k \in H , \forall t \in P \label{eq-15} \\
	&&& x_{ikt'}, y_{ikt}, y'_{ikt} \in \{0,1,2\},  \forall i \in M, \forall k \in H, \forall t \in  P, \forall t'\in PExt \label{eq-16}  \\
	&&& u_{ijkt} \in N, \forall i \in MExt, \forall k \in H, \forall t \in PExt \label{eq-17} \\
	&&& w_t \in \{0,1\}, \forall t \in P \label{eq-18}\\
	&&& prd_{it} \in N, \forall i \in M, \forall t \in P  \label{eq-19}\\
	&&& z_{ijkt} \in \{0,1\}, \forall i \in MExt, \forall j \in M, \forall k \in H, \forall t \in PExt \label{eq-20}\\ \notag	
    \end{align}

The objective function (\ref{eq-1}) minimizes the makespan. As noted before, the delivery time and the most relevant production costs are minimized when makespan is minimized. Constraints (\ref{eq-2}) and (\ref{eq-3}) are formulated to avoid solutions with empty intermediate periods. Constraints (\ref{eq-4}) correspond to the capacity of each heater (at most two molds). Constraints (\ref{eq-5}) - (\ref{eq-7}) correspond to the production capacity of each heater during a period $t$, taking into account curing and configuration times. Constraints (\ref{eq-8}) state the production amount for each type of tire and constraints (\ref{eq-9}) ensure that the demands are satisfied. Constraints (\ref{eq-10}) relate the number of molds used in period $t$ and heater $k$ to the $z$ assignment variables.
Constraints (\ref{eq-11}) and (\ref{eq-12}) correspond to the number of molds and parts available. Constraints (\ref{eq-13}) define which molds are already present at the heaters in the first period. Constraints (\ref{eq-14}) and (\ref{eq-15}) correspond to the number of mold $i$ setups and removals in period $t$ and heater $k$. Last, constraints (\ref{eq-16}) - (\ref{eq-20}) define the domains of the variables. 

\section{Solution approaches}
\label{hybrid}

For small instances it is possible to directly employ the MILP formulation given in the previous section and use a commercial solver for finding an optimal solution of the problem.
It is relevant to  note that model (\ref{eq-1}) - (\ref{eq-20}) has $O(|M|^2\times|H|\times \text{THB})$ integer variables and $O((|M|^2 \times|H|+|Q|)\times \text{THB}$) constraints. Therefore, the $\text{THB}$ value has a significant impact on the size of the model. Results from \citet{CPV2018} show that when $\text{THB}$ is large, solution times grow and  the solvers can not reach optimality and sometimes not even a feasible solution. On the other hand, if $\text{THB}$ is too low, there may not exist a feasible solution to the problem. 

The following formula allows to determine a $\text{THB}$ value that guarantees feasibility:  

{\small
\mathCfg
\begin{align}
	X = & \{i\in M : nm_i \ge 2 \wedge rq_{iq}=0, \forall q \in Q \} \notag \\
	Y = & \{i\in M : nm_i = 1 \vee  rq_{iq}=1, \exists q \in Q \} \notag  \\
	{tv}_i = & \max_{k\in H} \{ tv_{i,k} \}, \forall i \in M \notag  \\
	\text{THB} = & 
	\sum_{ i\in X} \Bigg\lceil \frac{ 4\lceil tc_i/{tv}_i)\rceil + 4\lceil tq_i/{tv}_i)\rceil  + dm_i}{2\lfloor \phi/{tv}_i\rfloor}\Bigg\rceil
	+
	\sum_{ i\in Y} \Bigg\lceil\frac{ \lceil tc_i/{tv}_i)\rceil + \lceil tq_i/{tv}_i)\rceil  + dm_i}{\lfloor \phi/{tv}_i\rfloor}\Bigg\rceil
	\label{improvedEstimator}
\end{align}	
}

Formula (\ref{improvedEstimator}) corresponds to the makespan value for producing sequentially all the tires in a single heater. Two sets of tires are considered: $X$ for those tires for which there are at least two identical molds and no parts are required, and $Y$ for tires with only one existing mold or requiring parts. 

We note that the $\text{THB}$ of (\ref{improvedEstimator}) can be much larger than the optimal makespan of a real situation in which there is more than one heater. Below we describe a heuristic procedure suggested for obtaining a feasible tight value for the $\text{THB}$ value, which when substituted in the  MILP formulation above discussed allows to reduce its size and therefore the solving time required. 

\subsection{Heuristic procedure}
\label{heuristic}

The heuristic procedure consists of a constructive phase and an improvement phase, embedded in an iterative search for producing new solutions for the problem. From all the solutions constructed, the one with the lowest makespan value is returned. We present below the main ideas of the procedure. For the readers interested in the details of the heuristic a pseudocode is provided in 
\ref{pseudocode}.

In the first step of the constructive phase, a set of pairs of molds is randomly determined, considering the available number of molds and shared parts between them as well as the combinations allowed. The production quantity of each pair of molds is also determined considering the demand requirements for each type of tire. The second step of the constructive phase builds a feasible solution for the problem by assigning pairs of molds to heaters, taking into account processing and configuration times as well as heaters availability. 

The improvement phase attempts to decrease the makespan value of the feasible solution obtained from the constructive phase by means of balancing as well as reducing the load of the different heaters. First, the assignments of pairs of molds without the empty mold are considered for splitting. 
Secondly, the excess in the heaters load is revised in order to reduce it.
The improvement phase finishes when a certain maximum number of iterations has been reached.

Although the order of complexity of the heuristic procedure described above is pseudopolynomial (see \ref{pseudocode}  for details), we want to note that it showed very fast execution times in the numerical experiments carried out and reported in Section \ref{experiments}.

\subsection{Hybrid optimization procedure}
\label{hybrid-procedure}
The hybrid optimization procedure (HOP) suggested for the tire scheduling problem stated at the beginning of Section \ref{form} consists of solving the MILP formulation of (\ref{eq-1}) - (\ref{eq-20}), using as $\text{THB}$ value the makespan of a given feasible solution. The feasible solution is that obtained from the heuristic procedure of Section \ref{hybrid}. First, the heuristic procedure is executed for obtaining an initial feasible solution of the problem, which in addition is considered as candidate for global solution. The makespan of the candidate solution is then set as the value for the $\text{THB}$ parameter of the MILP, and a MILP solver is invoked for obtaining an optimized solution, if it is possible. The procedure returns the solution with the minimum makespan obtained. 

\noindent \rule{67mm}{0.1mm}\\
The hybrid optimization procedure (HOP):\\
\noindent \rule{67mm}{0.1mm}

\begin{program}
\BEGIN
    candidateSolution := |Heuristic_Procedure|;
    globalSolution \xleftarrow{} candidateSolution;
    \text{THB}=candidateSolution.|getMakespan|();
    improvedSolution := |MILP_Solver|(\text{THB});
    \IF (improvedSolution.|getMakespan|() < globalSolution.|getMakespan|()) \THEN
            globalSolution \xleftarrow{} improvedSolution;
    \END 
    |return | globalSolution;
\END
\label{hybrid_pseudocode}
\end{program}



\section{Numerical experimentation}\label{experiments}

Here we present the numerical experiments carried out in order to evaluate the hybrid approach proposed for the tire curing scheduling problem at FunsaCoop. In order to perform this evaluation, the MILP formulation of Section~\ref{form} using the $\text{\text{THB}}$ of (\ref{improvedEstimator}) is used as a baseline. The results of this formulation are then compared against the hybrid optimization procedure (HOP) presented in Section \ref{hybrid-procedure}. 

\subsection{Instances description}

Three scenarios were defined in order to cover different sizes of instances as well as different degrees of resolution complexity. Each scenario is composed by fifteen instances generated randomly from
real life data provided by FunsaCoop.
The demand values for each type of tire were generated uniformly within an interval between $-20\%$ and $+20\%$ from a demand baseline according to the tire type and instance size scenario. From this procedure, in the case of the scenario of small instances S01 to S15, the demand values generated are in the range $[22, 595]$. For the scenario of medium size instances M01 to M15, the values are in the range $[111, 3012]$, and for the large instances L01 to L15, the range is $[1619, 7147]$. 
For all the scenarios, the number of molds type and heaters is in the set $\{5,7\}$ and $\{7,12\}$, respectively. The following sets of compatible molds with each other are considered: $\{1,...,5\}$, $\{6,7\}$ and $\{8,9\}$. The mold-heater combinations allowed are as follows: molds in $\{1,...,5\}$ with heaters in $\{1,..,7\}$; molds in $\{6,7\}$ with heaters in $\{8,9,10\}$; and molds in $\{8,9\}$ with heaters in $\{11,12\}$. The duration of a period is assumed equal to 1440 minutes (24 hours). The values of the time to place a mold in a heater are in the set $\{41.6, 60.6, 66.8\}$ (minutes) and the times to remove a mold from a heater are in the set $\{25.2, 44.9, 62.2\}$. The times considered for the tire curing are in the set $\{12.5, 18, 26, 30, 40, 42, 53, 55\}$ (minutes). A single unit per mold type is considered for the small instances; two units for medium size instances; and two, ten or fifteen units for the large instances. Two different mold shared parts are considered. For all the scenarios, molds 1 and 2 share a single unit of part 1. In addition, for the large instances molds 3 and 4 share two units of a single unit of part 2. 

The full instances, as well as other useful information, are available at \url{https://www.fing.edu.uy/inco/investigacion/proyecto/Funsacoop}.

\subsection{Computational results}

The model provided in Section \ref{form} was coded in AMPL and solved using CPLEX 12.6.0.0 with running time limited to 3600 seconds (1 hour). The heuristic procedure described in Section \ref{heuristic} was developed in Java 8. For managing  the execution of parallel threads, the library Threadly (\url{https://github.com/threadly/threadly}) was used. The number of total iterations was set to 100, 250 and 500, for small, medium and large instances, respectively. All the numerical experiments were executed in a laptop Intel Quad-Core 3.0Ghz with 16GB of RAM and Windows 10.


Results are reported in Tables \ref{small-cases}, \ref{medium-cases} and \ref{big-cases}. The first column of the tables shows the name of the instance. The second column shows the time horizon bound, $\text{THB}$, computed by means of formula \eqref{improvedEstimator} discussed at the end of Section \ref{form}. Columns three to five correspond to the makespan value, the duality gap and the execution time obtained from CPLEX solver for the MILP of Section~\ref{form} using the $\text{THB}$ of \eqref{improvedEstimator}. Column six shows the makespan value obtained with the heuristic procedure proposed in Section \ref{heuristic}. Finally, columns seven to nine correspond to the makespan value, the duality gap and the execution time obtained from the hybrid optimization procedure of Section~\ref{hybrid-procedure}. The running time of the heuristic procedure is not reported since it was less than 2 seconds for all instances. In addition, we point out that the cases where the duality gap value of columns four and eight is positive and the running time reported is less than the imposed limit of 3600 seconds correspond to situations where the CPLEX solver finished prematurely due to out-of-memory errors. 

A relevant information is the size of the MILP instances, which are reported in Tables \ref{size-small-cases}, \ref{size-medium-cases} and \ref{size-big-cases}. The tables include the name of the instances,  the $\text{THB}$ values, and the number of constraints, binary variables and real variables, both for the  MILP formulation with the initial $\text{THB}$ value and for the hybrid procedure. 

\begin{table}[ht!]
\begin{tabular}{|ccccccccc|}
\hline
\multicolumn{1}{|c}{\multirow{2}{*}{Instance}}  
& \multicolumn{4}{|c|}{MILP} 
& \multicolumn{4}{c|}{HOP} \\ 
\cline{2-5} \cline{6-9} 
& \multicolumn{1}{|c|}{\text{THB}} & \multicolumn{1}{c|}{Makespan} & \multicolumn{1}{c|}{\% Gap} & \multicolumn{1}{c|}{Time (s)} & \multicolumn{1}{c|}{\text{THB}} & \multicolumn{1}{c|}{Makespan} & \multicolumn{1}{c|}{\% Gap} & Time (s) \\ \hline
S01  &  17  &  8  &  0  &  2.870  &  8  &  8  &  0  &  1.141 \\ \hline
S02  &  20  &  9  &  0  &  1.201  &  9  &  9  &  0  &  0.456 \\ \hline
S03  &  21  &  9  &  0  &  1.307  &  9  &  9  &  0  &  0.492 \\ \hline
S04  &  19  &  8  &  0  &  1.628  &  8  &  8  &  0  &  0.448 \\ \hline
S05  &  19  &  8  &  0  &  1.464  &  8  &  8  &  0  &  0.469 \\ \hline
S06  &  44  &  12  &  0  &  5.174  &  12  &  12  &  0  &  0.989 \\ \hline
S07  &  43  &  15  &  0  &  6.352  &  15  &  15  &  0  &  1.003 \\ \hline
S08  &  44  &  12  &  0  &  6.591  &  13  &  12  &  0  &  1.101 \\ \hline
S09  &  45  &  13  &  0  &  5.724  &  13  &  13  &  0  &  0.900 \\ \hline
S10  &  43  &  14  &  0  &  4.756  &  14  &  14  &  0  &  0.616 \\ \hline
S11  &  44  &  13  &  0  &  5.054  &  13  &  13  &  0  &  0.621 \\ \hline
S12  &  48  &  17  &  0  &  5.057  &  17  &  17  &  0  &  1.023 \\ \hline
S13  &  46  &  17  &  0  &  4.867  &  17  &  17  &  0  &  1.198 \\ \hline
S14  &  45  &  16  &  0  &  4.123  &  16  &  16  &  0  &  0.746 \\ \hline
S15  &  39  &  13  &  0  &  3.320  &  13  &  13  &  0  &  0.625 \\ \hline
Average  &    &    &  0  &  3.966  &    &    &  0  &  0.789 \\ \hline
\end{tabular}
\caption{Makespan results for the small-size instances.}
\label{small-cases}
\end{table}

\begin{table}[ht!]
\begin{tabular}{|ccccccccc|}
\hline
\multicolumn{1}{|c}{\multirow{2}{*}{Instance}}  
& \multicolumn{4}{|c|}{MILP} 
& \multicolumn{4}{c|}{HOP} \\ 
\cline{2-5} \cline{6-9} 
& \multicolumn{1}{|c|}{\text{THB}} & \multicolumn{1}{c|}{Constraints} & \multicolumn{1}{c|}{Binary v.} & \multicolumn{1}{c|}{Real v.} & \multicolumn{1}{c|}{\text{THB}} & \multicolumn{1}{c|}{Constraints} & \multicolumn{1}{c|}{Binary v.} & Real v. \\ \hline
S01  &  17  &  4850  &  3397  &  1456  &  8  &  2150  &  1561  &  637 \\ \hline
S02  &  20  &  5750  &  4009  &  1729  &  9  &  2450  &  1765  &  728 \\ \hline
S03  &  21  &  6050  &  4213  &  1820  &  9  &  2450  &  1765  &  728 \\ \hline
S04  &  19  &  5450  &  3805  &  1638  &  8  &  2150  &  1561  &  637 \\ \hline
S05  &  19  &  5450  &  3805  &  1638  &  8  &  2150  &  1561  &  637 \\ \hline
S06  &  44  &  16582  &  11171  &  4859  &  12  &  4294  &  2979  &  1243 \\ \hline
S07  &  43  &  16198  &  10915  &  4746  &  15  &  5446  &  3747  &  1582 \\ \hline
S08  &  44  &  16582  &  11171  &  4859  &  13  &  4678  &  3235  &  1356 \\ \hline
S09  &  45  &  16966  &  11427  &  4972  &  13  &  4678  &  3235  &  1356 \\ \hline
S10  &  43  &  16198  &  10915  &  4746  &  14  &  5062  &  3491  &  1469 \\ \hline
S11  &  44  &  15936  &  10863  &  4558  &  13  &  4497  &  3144  &  1272 \\ \hline
S12  &  48  &  17410  &  11858  &  4982  &  17  &  5973  &  4140  &  1696 \\ \hline
S13  &  46  &  16672  &  11360  &  4770  &  17  &  5971  &  4139  &  1696 \\ \hline
S14  &  45  &  16305  &  11112  &  4664  &  16  &  5604  &  3891  &  1590 \\ \hline
S15  &  39  &  14091  &  9618  &  4028  &  13  &  4497  &  3144  &  1272 \\ \hline
Average  &    &  12699  &  8643  &  3698  &    &  4137  &  2891  &  1193 \\ \hline
\end{tabular}
\caption{Problem size data for the small-size instances.}
\label{size-small-cases}
\end{table}

\begin{table}[ht!]
\begin{tabular}{|ccccccccc|}
\hline
\multicolumn{1}{|c}{\multirow{2}{*}{Instance}}  
& \multicolumn{4}{|c|}{MILP} 
& \multicolumn{4}{c|}{HOP} \\ 
\cline{2-5} \cline{6-9} 
& \multicolumn{1}{|c|}{\text{THB}} & \multicolumn{1}{c|}{Makespan} & \multicolumn{1}{c|}{\% Gap} & \multicolumn{1}{c|}{Time (s)} & \multicolumn{1}{c|}{\text{THB}} & \multicolumn{1}{c|}{Makespan} & \multicolumn{1}{c|}{\% Gap} & Time (s) \\ \hline
M01  &  56  &  22  &  0  &  60.535  &  22  &  22  &  0  &  1.802 \\ \hline
M02  &  53  &  22  &  0  &  25.488  &  22  &  22  &  0  &  1.691 \\ \hline 
M03  &  58  &  28  &  0  &  33.615  &  28  &  28  &  0  &  1.840 \\ \hline
M04  &  52  &  20  &  0  &  18.837  &  20  &  20  &  0  &  1.061 \\ \hline
M05  &  59  &  27  &  0  &  42.493  &  27  &  27  &  0  &  1.910 \\ \hline
M06  &  201  &  71  &  0  &  146.778  &  72  &  71  &  0  &  7.667 \\ \hline
M07  &  198  &  60  &  0  &  164.959  &  61  &  60  &  0  &  7.197 \\ \hline
M08  &  187  &  64  &  0  &  189.438  &  65  &  64  &  0  &  6.996 \\ \hline
M09  &  212  &  77  &  0  &  147.855  &  76  &  76  &  0  &  34.512 \\ \hline
M10  &  208  &  85  &  1.18  &  3600.318  &  85  &  84  &  0  &  43.535 \\ \hline
M11  &  118  &  37  &  0  &  77.408  &  37  &  37  &  0  &  3.718 \\ \hline
M12  &  101  &  30  &  0  &  50.634  &  30  &  30  &  0  &  2.376 \\ \hline
M13  &  106  &  31  &  0  &  48.675  &  31  &  31  &  0  &  2.568 \\ \hline
M14  &  115  &  38  &  0  &  69.224  &  38  &  38  &  0  &  3.253 \\ \hline
M15  &  114  &  38  &  0  &  57.243  &  38  &  38  &  0  &  3.696 \\ \hline
Average  &    &    &  0.079  &  315.567  &    &    &  0  &  9.874 \\ \hline
\end{tabular}
\caption{Makespan results for the medium-size instances.}
\label{medium-cases}
\end{table}

\begin{table}[ht!]
\begin{tabular}{|ccccccccc|}
\hline
\multicolumn{1}{|c}{\multirow{2}{*}{Instance}}  
& \multicolumn{4}{|c|}{MILP} 
& \multicolumn{4}{c|}{HOP} \\ 
\cline{2-5} \cline{6-9} 
& \multicolumn{1}{|c|}{\text{THB}} & \multicolumn{1}{c|}{Constraints} & \multicolumn{1}{c|}{Binary v.} & \multicolumn{1}{c|}{Real v.} & \multicolumn{1}{c|}{\text{THB}} & \multicolumn{1}{c|}{Constraints} & \multicolumn{1}{c|}{Binary v.} & Real v. \\ \hline
M01  &  56  &  19798  &  9043  &  10115  &  22  &  7694  &  3535  &  3927 \\ \hline
M02  &  53  &  18730  &  8557  &  9569  &  22  &  7694  &  3535  &  3927 \\ \hline
M03  &  58  &  20508  &  9366  &  10479  &  28  &  9828  &  4506  &  5019 \\ \hline
M04  &  52  &  18374  &  8395  &  9387  &  20  &  6982  &  3211  &  3563 \\ \hline
M05  &  59  &  20864  &  9528  &  10661  &  27  &  9472  &  4344  &  4837 \\ \hline
M06  &  201  &  76649  &  49946  &  22600  &  72  &  27242  &  17825  &  8023 \\ \hline
M07  &  198  &  75504  &  49201  &  22261  &  61  &  23033  &  15088  &  6780 \\ \hline
M08  &  187  &  71289  &  46461  &  21018  &  65  &  24563  &  16083  &  7232 \\ \hline
M09  &  212  &  80860  &  52684  &  23843  &  76  &  28772  &  18820  &  8475 \\ \hline
M10  &  208  &  79326  &  51687  &  23391  &  85  &  32217  &  21060  &  9492 \\ \hline
M11  &  118  &  52324  &  22031  &  27859  &  37  &  16279  &  6884  &  8662 \\ \hline
M12  &  101  &  44761  &  18853  &  23830  &  30  &  13166  &  5576  &  7003 \\ \hline
M13  &  106  &  46986  &  19788  &  25015  &  31  &  13611  &  5763  &  7240 \\ \hline
M14  &  115  &  50989  &  21470  &  27148  &  38  &  16724  &  7071  &  8899 \\ \hline
M15  &  114  &  50544  &  21283  &  26911  &  38  &  16724  &  7071  &  8899 \\ \hline
Average  &    &  48500  &  26553  &  19606  &    &  16933  &  9358  &  6799 \\ \hline
\end{tabular}
\caption{Problem size data for the medium-size instances.}
\label{size-medium-cases}
\end{table}

\begin{table}[ht!]
\begin{tabular}{|ccccccccc|}
\hline
\multicolumn{1}{|c}{\multirow{2}{*}{Instance}}  
& \multicolumn{4}{|c|}{MILP} 
& \multicolumn{4}{c|}{HOP} \\ 
\cline{2-5} \cline{6-9} 
& \multicolumn{1}{|c|}{\text{THB}} & \multicolumn{1}{c|}{Makespan} & \multicolumn{1}{c|}{\% Gap} & \multicolumn{1}{c|}{Time (s)} & \multicolumn{1}{c|}{\text{THB}} & \multicolumn{1}{c|}{Makespan} & \multicolumn{1}{c|}{\% Gap} & Time (s) \\ \hline
L01  &  136  &  41  &  2.44  &  3600.198  &  40  &  40  &  0  &  5.531 \\ \hline
L02  &  152  &  49  &  0  &  2211.056  &  49  &  49  &  0  &  5.765 \\ \hline
L03  &  159  &  50  &  0  &  3180.421  &  50  &  50  &  0  &  5.614 \\ \hline
L04  &  145  &  51  &  1.96  &  3614.256  &  50  &  50  &  0  &  4.160 \\ \hline
L05  &  135  &  38  &  2.63  &  3600.191  &  37  &  37  &  0  &  2394.454 \\ \hline
L06  &  206  &  38  &  0  &  137.844  &  46  &  38  &  0  &  27.022 \\ \hline
L07  &  226  &  36  &  0  &  296.702  &  45  &  36  &  0  &  329.915 \\ \hline
L08  &  216  &  37  &  2.7  &  3600.412  &  44  &  37  &  2.7  &  3600.110 \\ \hline
L09  &  220  &  35  &  0  &  443.196  &  44  &  35  &  0  &  81.961 \\ \hline
L10  &  235  &  42  &  0  &  121.713  &  48  &  42  &  0  &  7.457 \\ \hline
L11  &  226  &  46  &  2.17  &  3600.307  &  49  &  46  &  2.17  &  3608.251 \\ \hline
L12  &  215  &  41  &  2.44  &  3608.348  &  48  &  41  &  2.44  &  3606.455 \\ \hline
L13  &  222  &  37  &  0  &  282.235  &  43  &  37  &  0  &  17.073 \\ \hline
L14  &  217  &  43  &  2.33  &  3600.384  &  42  &  42  &  0  &  3608.318 \\ \hline
L15  &  234  &  43  &  0  &  401.239  &  55  &  43  &  0  &  96.069 \\ \hline
Average  &    &    &  1.111  &  2153.233  &    &    &  0.487  &  1159.877 \\ \hline
\end{tabular}
\caption{Makespan results for the large instances.}
\label{big-cases}
\end{table}

\begin{table}[ht!]
\begin{tabular}{|ccccccccc|}
\hline
\multicolumn{1}{|c}{\multirow{2}{*}{Instance}}  
& \multicolumn{4}{|c|}{MILP} 
& \multicolumn{4}{c|}{HOP} \\ 
\cline{2-5} \cline{6-9} 
& \multicolumn{1}{|c|}{\text{THB}} & \multicolumn{1}{c|}{Constraints} & \multicolumn{1}{c|}{Binary v.} & \multicolumn{1}{c|}{Real v.} & \multicolumn{1}{c|}{\text{THB}} & \multicolumn{1}{c|}{Constraints} & \multicolumn{1}{c|}{Binary v.} & Real v. \\ \hline
L01  &  136  &  54124  &  19174  &  33215  &  40  &  15820  &  5638  &  9695 \\ \hline
L02  &  152  &  60506  &  21429  &  37135  &  49  &  19409  &  6906  &  11900 \\ \hline
L03  &  159  &  63299  &  22416  &  38850  &  50  &  19808  &  7047  &  12145 \\ \hline
L04  &  145  &  57713  &  20442  &  35420  &  50  &  19808  &  7047  &  12145 \\ \hline
L05  &  135  &  53725  &  19033  &  32970  &  37  &  14623  &  5215  &  8960 \\ \hline
L06  &  206  &  90656  &  38485  &  48715  &  46  &  20096  &  8565  &  10795 \\ \hline
L07  &  226  &  99474  &  42224  &  53455  &  45  &  19653  &  8377  &  10558 \\ \hline
L08  &  216  &  95064  &  40354  &  51085  &  44  &  19212  &  8190  &  10321 \\ \hline
L09  &  220  &  96828  &  41102  &  52033  &  44  &  19212  &  8190  &  10321 \\ \hline
L10  &  235  &  103439  &  43905  &  55588  &  48  &  20972  &  8936  &  11269 \\ \hline
L11  &  226  &  99008  &  42224  &  53448  &  49  &  21305  &  9125  &  11499 \\ \hline
L12  &  215  &  94181  &  40168  &  50841  &  48  &  20868  &  8939  &  11262 \\ \hline
L13  &  222  &  97252  &  41476  &  52500  &  43  &  18671  &  8003  &  10077 \\ \hline
L14  &  217  &  95057  &  40541  &  51315  &  42  &  18232  &  7816  &  9840 \\ \hline
L15  &  234  &  102518  &  43719  &  55344  &  55  &  23937  &  10246  &  12921 \\ \hline
Average  &    &  84190  &  34446  &  46794  &    &  19442  &  7883  &  10914 \\ \hline
\end{tabular}
\caption{Problem size data for the large instances.}
\label{size-big-cases}
\end{table}



From Table (\ref{small-cases}), it can be noted that for the scenario of small instances, both the MILP approach and the hybrid one obtain the optimal makespan in a reasonable computational time. However, we note that on average the hybrid approach is almost five times faster than the basic MILP. 

In the case of medium size instances the results of Table (\ref{medium-cases}) show that the basic MILP achieves the optimal makespan in 14 of 15 instances. We note that the hybrid approach is able to optimally solve all instances, and that a much lower  running time. 

For the large instances, the results of Table (\ref{big-cases}) show that the basic MILP approach achieves the optimal makespan for only 8 of 15 instances (53\%), whereas the hybrid approach achieves optimality for 11 of 15 instances (63\%). For the remaining four instances, we note that the duality gap of the hybrid approach is less than 2.7\%. 
Regarding the running times, we note that for these harder instances, the hybrid approach requires on average 54\% of the running time of the basic MILP approach. This comparison is not entirely accurate, since for several instances the running time reported for the basic MILP corresponds to the case of termination  due to the 3600 seconds CPLEX run time limit (which if not imposed, would lead to still longer solution times). For many instances where the basic MILP approach can not find the optimal value, the hybrid approach takes profit of the significantly reduced makespan values obtained by the heuristic, and builds a smaller model which can be solved to optimality. Nevertheless, we note that for some of the more demanding case the hybrid approach is not able to obtain a significantly improved result within the 3600 seconds running time limit. 

Considering the above results, we note that the proposed hybrid approach is able to obtain optimal solutions in the case of small and medium size instances of the tire curing scheduling problem under consideration, with a significntly improved computational performance over the basic MILP approach. In the case of large instances, the proposed hybrid approach also dominates the basic MILP approach. The hybrid approach obtains high quality solutions,  and in many cases optimal ones, including some instances where the basic MILP does not reach optimality. The efficiency is also increased, with lower execution times of the solver. 
From Tables (\ref{size-small-cases}), (\ref{size-medium-cases}) and (\ref{size-big-cases}), we can note that the number of constraints and variables is much lower in the hybrid approach than in the initial MILP formulation, as a smaller $\text{THB}$ value allows to have a more compact formulation. 
Therefore, the hybrid approach can be applied successfully to the problem in general, and in particular to the case of large instances.

\section{Conclusions}\label{conclusions}

This paper proposes and evaluates a hybrid optimization procedure for a tire curing scheduling problem from an Uruguayan cooperative tire factory. The tire curing is the bottleneck of the production line of the firm, and has an important impact in costs and energy consumption. This is a complex problem, since a limited number of resources as well as different tasks duration must be considered. Nowadays, the scheduling is done by experienced workers with no computational aids. The hybrid approach consists of solving a MILP formulation for the problem using the makespan returned by a heuristic procedure as the planning horizon value. The heuristic is composed by a constructive phase followed by an improvement phase, both embedded in an iterative process in order to find a solution that minimizes the makespan. The MILP formulation introduced in \citet{CPV2018} and revised here 
minimizes the makespan, considering features of the curing process under consideration such as allowed combinations of mold-mold and mold-heater, setup and removal times for each type of mold, and limited number of parts, molds and heaters. In order to evaluate the hybrid approach, an extensive numerical experimentation was conducted with instances of different size and based on data from a real case. From the results of the numerical experiments carried out, it can be observed that the hybrid approach is able to find the minimum makespan for most the instances, even the large ones. This is largely due to the fact that the makespan returned by the heuristic allows to significantly reduce the size of the MILP formulation, and to solve it in a reasonable amount of time,  achieving optimality for most of the instances. This effect is especially important in the case of large instances, where a non-tight value for the planning horizon can result in an unsolvable model in practice. 

Future work includes working closely with Funsacoop in order to implement the resolution methods developed and presented here. In this sense, a software including a practical interface for entering the information and visualizing the scheduling results was developed for the suggested heuristic. Some preliminary experiments carried out show the superiority of the proposed methods against manual made schedules. However, we note that further interaction is needed in order to fully validate the software and implement it in the firm.  
Despite the good performance observed for the hybrid optimization procedure, it would be interesting to evaluate the impact of using other heuristic procedures for estimating a tight feasible makespan value. Regarding the problem, a possible direction for future research is to study its complexity classification, and to analyze the current MILP formulation in order to 
explore alternative formulations which could be solved more efficiently. It would be also interesting to extend the work developed here in order to include other particular characteristics of the tire production line at FunsaCoop such as shifts, availability of workers, or distances among heaters, mold depots and storage areas. More in general, it would be also interesting to include other processes of the tire production line, as discussed by \citet{Oulamara2009} for the building stage, or explore opportunities to consider similar production problems involving batch processing of in-progress products, as in \citet{MottaToledo2016} for a glass container industry and \citet{PerezPerales2016} for a firm of ceramic products.

\section*{Acknowledgments}
We thank the staff of FunsaCoop for their time and willingness to provide information about their productive process and on real data for the settings of the test scenarios. We also thank Agust\'in Guerra (Unidad de Extensi\'on, FING) for his help in establishing the contact with FunsaCoop, and Agust\'in Ghioldi and Sof\'ia Lemes who participated in the initial contacts and developed a first mathematical model for the tire curing process.

\section*{References}
\bibliographystyle{plainnat}
\bibliography{mybibfile}

\newpage
\appendix

\clearpage
\newpage
\appendix
\section{Pseudocode of the heuristic procedure}
\label{pseudocode}
Here we provide the pseudocode of the heuristic procedure described in Section \ref{heuristic} of the paper, for the tire curing scheduling problem under consideration. 

The heuristic receives as input the data of a problem instance $(I)$ including the demand requirements for each type of tire, the mold-mold and mold-heater combinations allowed, number of parts, molds and heaters as well as configuration and curing times. It also receives the value ofthe  total number of iterations (TOTAL\_ITERATIONS). The heuristic returns a solution for the problem specified as a collection of 5-tuples, where each 5-tuple includes: 1) a numerical identifier of the tuple; 2) a pairs of molds $(M1, M2)$; 3) the number of units  to produce of each mold $(Q)$; 3) the assigned heater $(H)$; 4) the starting period $(T)$, and 5) the number of periods of the assignment $(L)$. The demand requirements of the empty mold are assumed zero and the makespan value of an empty solution equal to infinite.

\noindent \rule{50mm}{0.1mm}\\
Heuristic main procedure:\\
\noindent \rule{50mm}{0.1mm}

\begin{program}
\BEGIN
    globalSolution := \emptyset;
    \FOR i := 1 | to TOTAL_ITERATIONS | \DO
        partialSolution := |mold_pairsProcedure()|;
        initialSolution := |assignmentProcedure(|partialSolution|)|;
        improvedSolution := |improvementProcedure(|initialSolution|)|;
        \IF (improvedSolution.|getMakespan|() < globalSolution.|getMakespan|()) \THEN
            globalSolution \xleftarrow{} improvedSolution;
        \END
        i := i+1;
    \END 
    |return | globalSolution;
\END
\end{program}

\noindent \rule{50mm}{0.1mm}\\
Auxiliary procedures:\\
\noindent \rule{50mm}{0.1mm}

\begin{program}
    \PROC |mold_pairsProcedure|() \BODY
        \FOR m := 1| to |I.|getNumberOfMoldTypes()| \DO
            Demand(m) := I.|getDemand|(m);
            \IF (I.|getQuantityOfMold|(m)>0) \THEN
                BatchSize(m) := I.|getDemand|(m)/I.|getQuantityOfMold|(m);
            \ELSE
                BatchSize(m) := \infty;
            \FI  
        \END  
        j := 1;
        solution := \emptyset;
        \WHILE (\exists t : Demand(t) > 0) \DO
            (M1,M2) := I.|getRandomMoldPairWithPositiveDemand|();
            \IF (|there are sufficient molds and parts for producing |M1| and |M2) \THEN
                \IF (M1 > 0| AND |M2 > 0) \THEN
                    Q:= min\{ BatchSize(M1), BatchSize(M2), Demand(M1), Demand(M2) \};
                \ELSE
                    \IF (M1 > 0) \THEN
                        Q:= min\{ BatchSize(M1), Demand(M1) \};
                    \ELSE
                        \IF (M2 > 0) \THEN
                            Q:= min\{ BatchSize(M2), Demand(M2) \};
                        \END
                    \END
                \END
                H := \emptyset;
                T:= \infty;
                L := \infty;
                solution \xleftarrow{} solution \cup \{(j, M1, M2, Q, H, T, L)\};
                Demand(M1) := Demand(M1) - Q;
                Demand(M2) := Demand(M2) - Q;
                j := j+1;
            \END                
        \END
        |return | solution;
    \END

    \PROC |assignmentProcedure|(partialSolution) \BODY 
        S \xleftarrow{} partialSolution;
        solution := \emptyset;
        \WHILE  S \neq \emptyset \DO
            \FOR | all | x \in S \DO
                x.T := solution.|getInitialPeriodOfAssignment(|x.M1,x.M2);
            \END
            p :=  solution.|getMoldsPairWithShortestStartTime|();
            p.H := solution.|getHeaterAvailableWithLowestSetups|(p.M1,p.M2);
            p.L := solution.|getNumberOfAssignementPeriods|(p.M1,p.M2,p.Q,p.H);
            solution \xleftarrow{} solution \cup \{p\};
            S \xleftarrow{} S|\textbackslash|\{p\};
        \END
        |return | solution;
    \END

    \PROC |improvementProcedure|(originalSolution) \BODY
        improvedSolution \xleftarrow{} originalSolution;
        improvement \xleftarrow{} \true;
        \WHILE (improvement) \DO
            S1 \xleftarrow{} improvedSolution;
            j := S1.|getLastNumericalIdentifier|();
            \FOR| all |x \in S1: (x.M1 \neq 0, x.M2 \neq 0) \DO
                H := \emptyset;
                T:= \infty;
                L := \infty;
                \IF x.M1 = x.M2 \THEN \\%
                    y1 := (j+1, x.M1, x.M2, (\lceil x.Q)/2 \rceil, H, T, L);
                    y2 := (j+2, x.M1, x.M2, (\lfloor x.Q)/2 \rfloor, H, T, L);
                \ELSE
                    y1 := (j+1, x.M1, 0, \lceil (x.Q)/2 \rceil, H, T, L);
                    y2 := (j+2, x.M2, 0, \lfloor (x.Q)/2 \rfloor, H, T, L);
                \END
                S1 \xleftarrow{} S1|\textbackslash|\{x1\}\cup \{y1,y2\};
                j := j+2;
            \END
            
            S2 := assignmentProcedure(S1);
            
            \FOR| all |h \in I.|getSetOfHeaters|() \DO
                x := S2.|getLastMoldPairOfHeater|(h);
                x.Q := S2.|tryReduceProduction|(x);
            \END
            
            \IF (S2.|getMakespan|() < improvedSolution.|getMakespan|()) \THEN
                improvedSolution \xleftarrow{} S2;
            \ELSE
                improvement \xleftarrow{} \false;
            \END
        
        \END
        |return | improvedSolution;
    \END

\end{program}

\noindent \rule{50mm}{0.1mm}\\
Other auxiliary procedures:\\
\noindent \rule{50mm}{0.1mm}

\begin{program}
    \PROC |getMakespan|() \BODY
        |Returns the makespan of a solution calculated as the maximal final period of|
        |completion |(T+L)| among all heaters|. 
    \END
    \PROC |getNumberOfMoldTypes|() \BODY
        |Returns the number of different mold types of a problem instance|.
    \END
    \PROC |getDemand|(m) \BODY
        |Returns the demand requirements for mold type |m| of a problem instance|.
    \END
    \PROC |getQuantityOfMold|(m) \BODY
        |Returns the quantity of molds for mold type |m| of a problem instance|.
    \END
    \PROC |getRandomMoldPairWithPositiveDemand|() \BODY
        |Returns a pair of compatible molds randomly selected |m1| and |m2
        |with positive net demand (demand value minus production quantity)|
        |for at least one of them|.
    \END
    \PROC |getInitialPeriodOfAssignment|(m1,m2) \BODY
        |Returns |max\{t1, t2, t3\}| with: |
        t1| the earliest period with compatible heater available for molds |m1| and |m2
        t2| the earliest period with available molds for molds |m1| and |m2
        t3| the earliest period with available parts for molds |m1| and |m2
    \END
    \PROC |getMoldsPairWithShortestStartTime|()
        |Returns the pair of molds with the shortest start time of a certain solution|.
    \END
    \PROC |getHeaterAvailableWithLowestSetups|(t,m1,m2) \BODY
        |Returns a compatible heater available at period |t| that requires|
        |less set-up times and removal times for |m1| and |m2.
    \END
    \PROC |getNumberOfAssignementPeriods|(h,q) \BODY
        |Returns the numbers of periods for producing a quantity |q
        |of molds m1 and m2 in heater |h.
    \END
    \PROC |getLastNumericalIdentifier|() \BODY
        |Returns the greatest numerical identifier value of the tuples of a certain solution|. 
    \END
    \PROC |getSetOfHeaters|() \BODY
        |Returns the set of heaters of a problem instance|.
    \END
    \PROC |getLastMoldPairOfHeater|(h) \BODY
        |Returns the last pair of molds assigned to heater |h| in a certain solution|.
    \END
    \PROC |tryReduceProduction|(x) \BODY
        |Returns a reduced production quantity for the pair of molds |(x.M1,x.M2)
        |whenever the current total production quantity for each|
        |mold is greater than the demand requirement|.
    \END
\end{program}
\noindent \rule{50mm}{0.1mm}\\

The complexity order of the heuristic procedure depends on the procedures for the assignment of pairs of molds to heaters (assignmentProcedure) and for improving the initial solution (improvementProcedure). The order of the assignment procedure is $O(D^2)$, with $D=\sum_{i\in M}dm_i$, the total demand requirements, since in the worst case a solution is composed by $O(D)$ tuples, each on of them with one unit of production quantity for a certain pair of molds. We note that all the procedures called in the assignment procedure run in $O(D)$ time. In the case of the improvement procedure, the order is $O(D^3)$ since for each tuple of a given solution it calls the assignment procedure. Therefore, the order of complexity of the heuristic procedure is $O(D^3)$.

\end{document}